\pdfoutput=1
\pdfoutput=1
\pdfoutput=1
\pdfoutput=1
\pdfoutput=1

\documentclass[journal]{IEEEtran}

\usepackage{graphicx}
\usepackage{epstopdf}
\usepackage{dcolumn}
\usepackage{bm}

\usepackage{float}
\usepackage{color}
\usepackage[caption=false]{subfig}

\usepackage[utf8]{inputenc}
\usepackage[T1]{fontenc}
\usepackage{mathptmx}

\usepackage{booktabs} 

\usepackage{algorithm}
\usepackage{algorithmic}

\usepackage{indentfirst}

\usepackage{mathptmx}

\usepackage[font=small,labelfont=bf,labelsep=period]{caption}

\begin{document}
%
\title{Real-time Human Activity Recognition Using Conditionally Parametrized Convolutions on Mobile and Wearable Devices}
%
%
%

\author{Xin Cheng, Lei Zhang, Yin Tang, Yue Liu, Hao Wu and Jun He,~\IEEEmembership{Member,~IEEE}

\thanks{This work was supported  in part by the National Science Foundation of China under Grant 61971228 and the Industry-Academia Cooperation Innovation Fund Projection of Jiangsu Province under Grant BY2016001-02, and in part by the Natural Science Foundation of Jiangsu Province under grant BK20191371.\textit{ (Corresponding author: Lei Zhang.)}}
\thanks{Xin Cheng, Lei Zhang, Yin Tang and Yue Liu are with School of Electrical and Automation Engineering, Nanjing Normal University, Nanjing, 210023, China (e-mail: leizhang@njnu.edu.cn).}
\thanks{Hao Wu is with School of Information Science and Engineering, Yunnan University, Kunming 650091, China.}

\thanks{Jun He is with School of Electronic and Information Engineering, Nanjing
	University of Information Science and Technology, Nanjing 210044, China.}

%

}


%



\maketitle

\begin{abstract}
Recently, deep learning has represented an important research trend in human activity recognition (HAR). In particular, deep convolutional neural networks (CNNs) have achieved state-of-the-art performance on various  HAR datasets. For deep learning, improvements in performance have to heavily rely on increasing model size or capacity to scale to larger and larger datasets, which inevitably leads to the increase of operations. A high number of operations in deep leaning increases computational cost and is not suitable for real-time HAR using mobile and wearable sensors. Though shallow learning techniques often are lightweight, they could not achieve good performance. Therefore, deep learning methods that can balance the trade-off between accuracy and computation cost is highly needed, which to our knowledge has seldom been researched. In this paper, we for the first time propose a computation efficient CNN using conditionally parametrized convolution for real-time HAR on mobile and wearable devices. We evaluate the proposed method on four public benchmark HAR datasets consisting of WISDM dataset, PAMAP2 dataset, UNIMIB-SHAR dataset, and OPPORTUNITY dataset, achieving state-of-the-art accuracy without compromising computation cost. Various ablation experiments are performed to show how such a network with large capacity is clearly preferable to baseline while requiring a similar amount of operations. The method can be used as a drop-in replacement for the existing deep HAR architectures and easily deployed onto mobile and wearable devices for real-time HAR applications.\\
\end{abstract}


%
\IEEEpeerreviewmaketitle

\section{Introduction}
%
%
%
%
\IEEEPARstart{H}{uman} activity recognition (HAR) has become an important research area in ubiquitous computing and human computer interaction, which has a variety of applications including health care, sports, interactive gaming, and monitoring systems for general purposes. With the rapid technical advancement of mobile phones and other wearable devices, various motion sensors have been placed at different body positions in order to collect data and infer human activity details \cite{bulling2014tutorial}. Unlike video or wireless signals based method, mobile phone and wearable devices are more popular, which are not location dependent, easy to deploy and have no any health hazard caused by radiation. As we have known, mobile phones have become an important part of human’s daily life and can be carried around almost every day. Therefore, the use of data generated by mobile phones and other wearable sensors has dominated the research landscape in HAR, which provides obvious advantages over other sensor modalities \cite{demrozi2020human}. On the whole, mobile and wearable sensor based methods provide a better alternative to real-time implementation of HAR applications  \cite{lara2012survey}.\\
\indent On the other hand, sensor based HAR mainly lies in the assumption that specific body movement can be translated into characteristic sensor signal pattern, which may be further classified using machine learning technique \cite{anguita2012human}. Recently, deep learning technique outperformed many conventional machine learning methods, which has represented an important research trend in HAR \cite{wang2019deep}. In particular, deep convolutional neural networks (CNNs) have achieved state-of-the-art performance on various HAR tasks \cite{yang2015deep}. For deep learning, improvements in performance have to heavily rely on increasing model size or capacity to scale to larger and larger datasets \cite{simonyan2014very}. However, increasing model size or capacity inevitably leads to the increase of operations or computation cost. Building larger CNN may result in higher performance, but lead to the need for more resources such as computational power that is expensive for mobile and wearable devices. Therefore, deploying optimal deep models for mobile and wearable HAR applications are often impractical, which limits their wide use for real-time HAR applications with strict latency constraints. Therefore, it deserves further research to develop computation efficient CNN to perform real-time HAR using mobile and wearable sensors.\\
\indent Without loss of generality, there is two ways to design computation efficient CNN for mobile and wearable HAR applications. For the first case, using fewer convolutional layers or decreasing the size of existing convolutions may lead to the decrease of computation cost. Thus, current computationally efficient models often are smaller, which have suboptimal performance with fewer parameters on mobile deployment \cite{howard2017mobilenets}, \cite{tang2020efficient}. For the second case, decreasing the size of the input to convolution also can proportionally decrease computation cost. Actually, HAR using mobile phones and wearable sensors can be seen as a classic multivariate time series classification problem, which makes use of sliding window \cite{banos2014window} to segment time series sensor signals and extracts discriminative features from them to be able to recognize activities by utilizing a classifier. Intuitively, using smaller sliding window can yield faster inference.    However, in this case it often is hard to obtain most suitable size for feature extraction of HAR, which make CNNs be not able to offer best results. Therefore, as indicated in both cases, current computationally efficient models often are suboptimal for HAR \cite{tang2020efficient}. Recently, there has been rising research interest in conditional computation \cite{bengio2013estimating},          \cite{cho2014exponentially}, whose goal is to increase model capacity or performance without a proportional increase in computation cost. In particular, Yang \emph{et~al.} \cite{yang2019condconv} proposed an idea of conditionally parameterized convolution (CondConv), which can easily be optimized by gradient descent. According to our research motivation, replacing conventional convolutions with CondConv could be one feasible step to realize efficient inference for mobile and wearable HAR applications without compromising computation cost.\\
\indent In this paper, we propose a new CNN using the idea of CondConv for HAR applications with strict latency constraints, which aims to increase model capacity or performance while maintaining efficient inference to better serve these real-time HAR applications on mobile and wearable devices. To the best of our     knowledge, building high-performance CNN for HAR without compromising computation cost has seldom been explored, and this paper is the first try to develop computation efficient CNN for real-time HAR on ubiquitous and wearable computing area. To be specific, we replace the standard convolution $\mathit{W*x}$ with CondConv which is a linear combination of  $\mathit{n}$ experts$\left(\left( \alpha_{1}\cdot W_{1} +\alpha_{2}\cdot W_{2}+...+\alpha_{n}\cdot W_{n} \right )* X\right)$, where $\alpha_{1}$, ...,$\alpha_{n}$ are weight functions of the input learned through gradient descent. The standard convolution  $\mathit{W*x}$ requires expensive computation cost as it needs to be computed at many different positions within the input. In comparison with standard convolution, increasing the number of experts in the CondConv can greatly enhance the representing ability of CNN without compromising computation cost, as all the experts are combined only once per input. Thus, we can increase model capacity or performance via increasing the number of experts, which require only an expensive convolution with a very small increase in computation cost. We evaluate our method on four public benchmark HAR datasets, namely WISDM dataset \cite{kwapisz2011activity}, PAMAP2 dataset \cite{reiss2012introducing}, UNIMIB-SHAR dataset \cite{micucci2017unimib}, and OPPORTUNITY dataset \cite{chavarriaga2013opportunity}. We also evaluate the actual inference speed of our model on a smartphone with an Android platform. By various ablation experiments, we show how increasing the number of experts improves model performance across the benchmark HAR datasets while maintaining efficient inference. The experimental results indicate the advantage of the HAR applications using CondConv with regard to typical challenges for real-time HAR in ubiquitous and wearable computing scenarios.\\
\indent The rest of this paper is organized as follows. Section II presents the related works in activity recognition and conditional computation. Section III details the proposed framework for HAR. In Section IV, we first describe the HAR dataset used and experimental setup, and then present the experimental result comparison and analysis from several aspects. The last section concludes this study with a brief summary and points out future research work.

\section{Related Works}
\indent In recent years, deep learning has become popular in mobile and wearable sensors based HAR, due to their superior performance. In particular, CNN is one of the most researched deep learning techniques which can automatically extract features and identify the hidden or unknown activity patterns from raw time series sensor data. A number of CNN architectures for the use of HAR have been developed by researchers. For example, Zeng \emph{et~al.} \cite{zeng2014convolutional} firstly proposed a shallow CNN based approach to recognize activities, which has achieved state-of-the-art performance in three public HAR datasets. Yang \emph{et~al.} \cite{yang2015deep} developed a new architecture of CNN, in which the convolution filters are applied along the temporal dimension for each sensor and all feature maps for different sensors are unified as an input for a classifier. CNNs that combine other fusion  techniques were also proposed. Ordóñez\emph{et~al.} \cite{article} proposed an architecture of DeepConvLSTM, which replaced the fully connected layer of CNN with Long Short Term Memory (LSTM) to capture temporal relationship contained in time series sensor data. Wang \emph{et~al.} \cite{wang2019attention}  proposed an attention-based CNN which is able to enhance interesting activity in the weakly supervised learning scenarios. Ignatov \emph{et~al.} \cite{ignatov2018real} proposed a CNN which combines local feature extraction with simple statistical features that preserve global information about the time series sensor data. Teng \emph{et~al.} \cite{TengQi} developed a layer wise training CNN for HAR with local loss, which is able to achieve remarkable performance with less parameters on various  HAR application domains. On the whole, deep CNNs have yielded excellent results in terms of recognition accuracy, but often need a lot of computation cost, which is infeasible for mobile and wearable HAR applications that have strict latency constraints.\\ 
\indent Due to the growing number of hyper-parameters, designing computation efficient CNN for HAR applications becomes increasingly difficult. In another line of research, recent research effort on visual recognition or natural language processing has been shifting to conditional computation, which aims to increase model capacity or performance without a proportional increase in computation cost. For example, Wu \emph{et~al.} \cite{Wu_2018_CVPR} proposed BlockDrop method which uses reinforcement learning to dynamically learn discrete routing functions, in order to best reduce computation cost without decreasing model accuracy. Mullapudi \emph{et~al.} \cite{teja2018hydranets} developed HydraNet model which uses unsupervised clustering method to choose proper subset of the entire network architecture to run most efficient inference on a given input. Shazeer \emph{et~al.} \cite{shazeer2017outrageously} proposed a trainable gating network by introducing a sparsely-gated mixture-of-experts layer, which is able to determine a sparse combination of different experts to use for each example. However, these aforementioned approaches in conditional computation often require to learn discrete routing decisions of different experts across every example, which is hard to train using gradient descent and not suitable for CNN based HAR applications. Recently, Yang \emph{et~al.} \cite{yang2019condconv} proposed CondConv to challenge the fundamental assumption that the same convolutional kernels should be shared for each example, which enables different expert convolution kernels to focus on their specialized examples. In particular, the CondConv can easily be trained with gradient descent without requiring access to discrete routing of each example. Despite the success of conditional computation, their primary use mainly lies in imagery or natural language processing tasks, which has never been used to perform HAR. The increasing demands for running efficient deep neural networks for HAR on mobile and wearable devices encourage our current study. In the next section, we will describe the CondConv and then present the entire architecture of deep HAR applications using CondConv.

\begin{figure}[htbp]
	\centering
	\hspace{-0.2cm}
	\includegraphics[height=6.8cm,width=8.4cm]{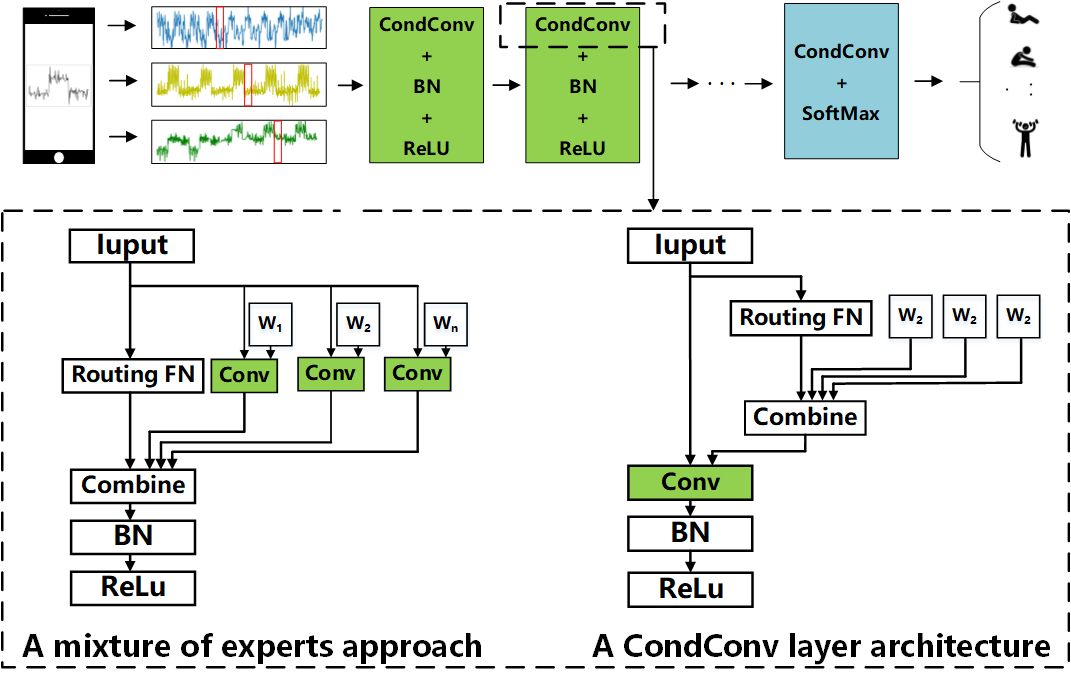}
	\caption{ Overview of the proposed HAR framework using CondConv}
	\label{2}
\end{figure}
\section{Proposed Scheme}

\indent In this section, we will discuss our new CNN architecture using CondConv to handle the unique challenges existed in mobile and wearable HAR applications. An overview of the proposed HAR system is presented in Fig.1. For sensor based HAR, we have to firstly deal with multiple channels of time series sensor signals, in which the convolution need to be applied along temporal dimension and then be shared or unified among multiple different sensors. Due to implementational simplicity and no need of preprocessing, the sliding window technique is ideally suitable for real-time HAR applications, which has been widely used to segment time series sensor signal into a collection of smaller data pieces as an input for CNN. Hence an instance handled by CNN typically corresponds to a two-dimensional matrix with r raw samples representing the number of samples per window, in which each sample contains multiple sensor attributes recorded at time t. Though in any case the sensor signal stream must be segmented into data windows, they can be of a continuous nature. Thus, an overlap between adjacent windows is tolerated to preserve the continuity of activities. Intuitively, decreasing the size of sliding window leads to a faster activity inference, as well as a reduced need for computation cost. To make fair comparison, we still select the same window size that is preferably used in previous state-of-the-art works. \\


\indent Our main research motivation is to realize computation efficient CNN using CondConv for the practical use of HAR on mobile and wearable devices. Without loss of generality, the baseline CNN is typically comprised of four units: (i) a convolution layer with a set of learned kernels that convolve the input along temporal dimension or the previous layer’s output; (ii) a ReLU layer with activation function max(x,0) that maps the previous layer’s output; (iii) a max pooling layer that subsamples via finding the maximum feature map across a range of local temporal neighborhood; (iv) a Batch Normalization(BN) \cite{ioffe2015batch} layer used to normalize the values of different feature maps from the previous layer. Following the settings of Yang \emph{et~al.} \cite{yang2019condconv}, we replace the standard convolution kernels used in convolutional layers with a linear combinations of $\mathit{n}$ experts:
\begin{equation}	
\mathit{Output}=\sigma\left(\left( \alpha_{1}\cdot W_{1} +...+\alpha_{n}\cdot W_{n} \right )* X\right)  
\end{equation}
 in which  $\sigma$ is  ReLU activation function and $\mathit{n}$ is the number of experts. The dimension of each kernel $W_i$ is still the same to that in original convolution. Obviously, if the scalar weight $\alpha$ is constant for all examples, a CondConv layer has almost the same capacity with a standard convolutional layer. To avoid the case, the weight $\alpha_{i}$  can be computed using a routing function $r_i$(x): 
\begin{equation}	
r_{i}(x)=S\left(\mathit{GlobalAveragePool\left(\mathit{x}\right)}*R\right)
\end{equation}
\noindent in which  $\mathit{S}$  is Sigmoid activation function, and GlobalAveragePool is global average pooling layer. $\mathit{R}$ is a dense layer that maps the pooled inputs to $\mathit{n}$ expert weights with the parameters learned across lots of training examples. Therefore, the weights of  $\mathit{n}$ experts are example-dependent, which enable different experts to specialize in their interesting examples. That is to say, the weights of $\mathit{n}$ experts are different across all examples, in which each individual example can be processed with different weights. \\
\indent From the perspective of matrix theory, a CondConv layer can be equally expressed as:
\begin{equation}	
\mathit{Output}=\sigma\left( \alpha_{1}\cdot \left(W_{1}*x\right)  +...+\alpha_{n}\cdot  \left(W_{n}*x\right)  \right)\\
\end{equation}
which is more computationally expensive. As a comparison, the CondConv for each example can be computed as a linear combination of $\mathit{n}$ experts, and then only one expensive convolution needs to be computed. To be specific, each additional expert requires only one additional multiply-add operation, which suggests that we can increase model capacity or performance via increasing the number of experts, with only a very small increase in computation cost. Though increasing the number of experts inevitably requires more memory resource, it is often affordable due to the rapid technical advancement of mobile phones and other wearable devices. Hence the CondConv is able to achieve higher inference performance without compromising computation cost, which provides a better alternative to serve mobile and wearable HAR that has strict latency constrains. With the increase in the number of experts, the CondConv is able to increase model capacity, which is also prone to overfitting. To avoid overfitting, we additionally introduce data augmentation via improving the overlapping rate of sliding windows, as well as randomly dropping out to ensure sufficient regularization.

\section{Experiments and Results}
\indent We evaluate the proposed method on four public benchmark HAR datasets consisting of WISDM dataset \cite{kwapisz2011activity}, PAMAP2 dataset \cite{reiss2012introducing}, UNIMIB-SHAR dataset \cite{micucci2017unimib}, and OPPORTUNITY dataset \cite{chavarriaga2013opportunity}, which are recorded with different sampling rates, number of sensors and kinds of activities. In terms of accuracy and FLOPs, we compare our method against the baseline CNN, as well as other state-of-the-art techniques that have been widely used in the HAR tasks. To make fair comparison, we restrain the baseline CNN with the same hyperparameters and regularization methods as the CondConv model. For each baseline architecture, we replace standard convolution layer with CondConv Layer to evaluate CondConv via increasing the number of experts per layer. To be specific, model performance is evaluated via varying the number of experts in the CondConv layer from {1, 2, 4, 8, 16}. To fully exert the effect of CondConv, we additionally replace the fully connected layer with a 1x1 CondConv layer in some cases. For each CondConv layer, the BN layer is inserted right after a convolutional layer, but before feeding into ReLU activation \cite{glorot2011deep}. To determine the routing weight functions, we experiment with different activation functions including Tanh, Sigmoid, Softmax, LReLU, ELU and ReLU, in which the results suggest that Sigmoid significantly outperforms other activation functions. Various ablation experiments are performed to further analyze the effect of CondConv layer across different examples at different depths in the network.\\

\indent Models are trained in a supervised way, and the model parameters are optimized by minimizing the cross-entropy loss function with mini-batch gradient descent using an Adam optimizer. Training is done for at least 400 epochs. The epoch that achieves the best performance is selected and the corresponding model is applied to test set. For the CondConv, increasing the number of experts will inevitably lead to the increase of parameter count, which requires enough examples to train the model. The data augmentation and dropout technique are used for the CondConv model with large capacity, which aims to ensure sufficient regularization. First, data augmentation technique is added via improving the overlapping rate of sliding time windows. We use smaller sliding step length to segment time series sensor signal, which is able to generate more training examples. Second, dropout technique is applied to avoid overfitting during the training stage. However, normal combination of dropout and BN technique often lead to worse results unless some conditioning is done to prevent the risk of variance shifts. As suggested by Li \emph{et~al.}  \cite{li2019understanding}, the worse performance caused by the variance shift only happens when there exists a dropout layer before a BN layer. Thus, we insert only one dropout layer right before the final Softmax layer. All the experiment in this paper are implemented in Python using TensorFlow backend on a machine with an Intel i7-6850K CPU, 64GB RAM and NVIDIA RTX 2080 Ti GPU. In addition, we test the actual inference speed on a smartphone with an Android platform.\\

\subsection{Experiment Results and Performance Comparison }

\subsubsection{$\textbf{The WISDM dataset}${\color{black} }  \cite{kwapisz2011activity}   }
\indent The WISDM dataset used for the experiment is provided by the Wireless Sensor Data Mining(WISDM) Lab, which contains various human activities with 6 attributes: user, activity, timestamp, x-acceleration, y-acceleration, z-acceleration. The smartphones were placed in a front leg pocket of each dominant, in which one triaxial accelerometer embedded in smartphones with an Android platform was used to generate time series data at a constant sampling rate of 20Hz. The activities were collected from 29 subjects and each subject performed 6 distinctive human activities consisting of walking, jogging, walking upstairs, walking downstairs, sitting and standing .\\
\indent In the experiment, the sliding window technique is utilized to segment the time series accelerometer signals. The size of sliding time window is set to 10s and a 95$\%$ overlapping rate is used, which equals to 0.5s of the sliding step length. The whole WISDM dataset is partitioned into two parts, in which 70$\%$ is randomly selected to generate training examples and the rest test examples. The shorthand description of the baseline CNN architecture is C(64)-C(128)-C(384)-FC-Sm, which consists of three convolutional layers and one fully connected layer. To be specific, each convolution begins with Conv-BN-ReLU and then another one. We use a 1x1 CondConv layer to replace the fully connected classification layer. The model will be trained using mini-batches with a size of 210. Adam is used for optimization. The initial learning rate is set as 0.0001, which will be reduced by a factor of 0.1 after each 50 epochs.

\begin{figure}[htbp]
	\centering
	\includegraphics[width=7.5cm,height=6cm]{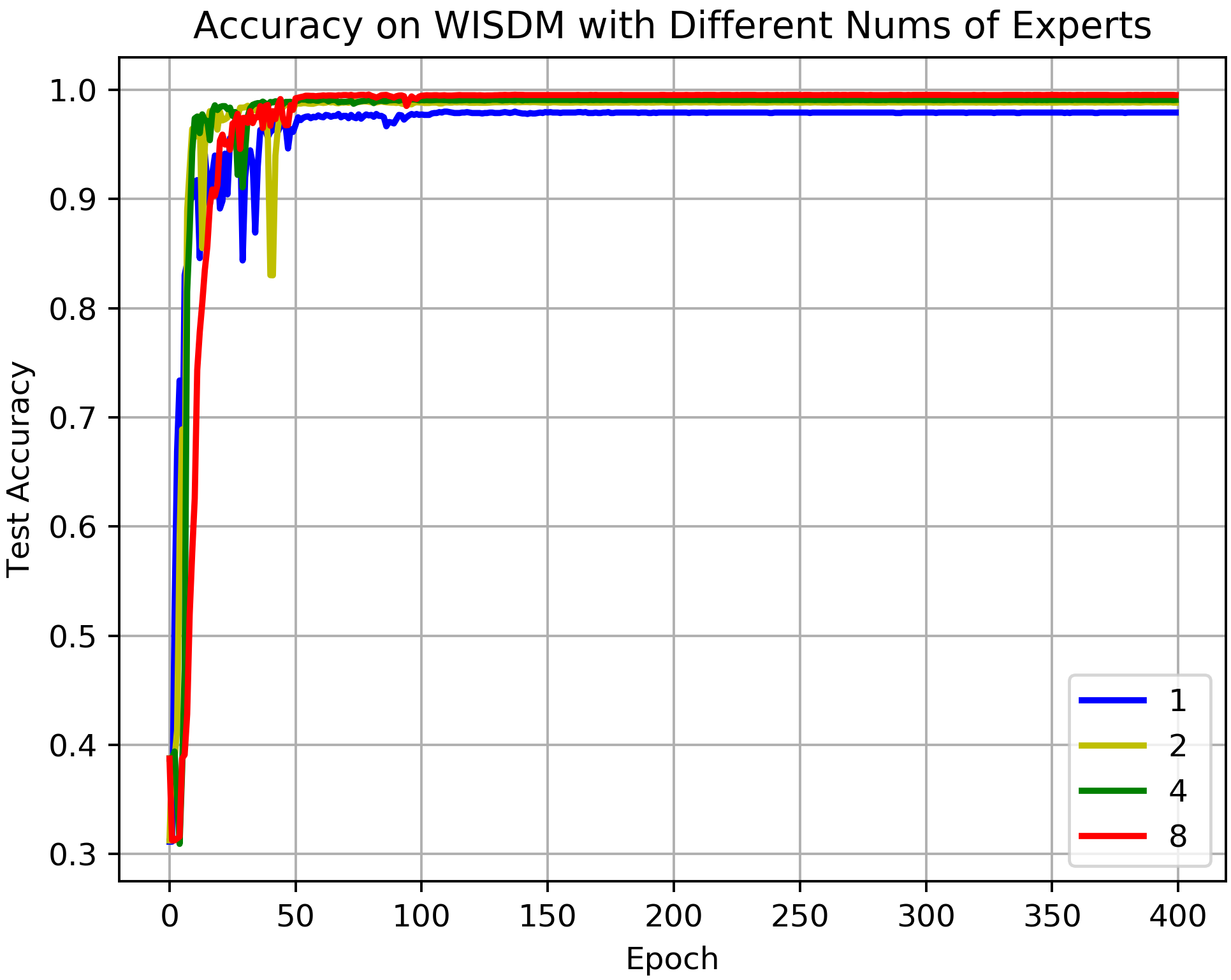}
	\caption{\label{Fig2 }  Accuracy on \textbf{WISDM} dataset with different nums of experts}
\end{figure}

\indent In Fig.2, we evaluate model performance using CondConv via varying the number of experts. As can be seen in the figure, there is a steady increase in performance on test data with increasing the number of experts. During training stage, increasing the number of experts makes the model converge faster. In terms of accuracy and FLOPs, Table \uppercase\expandafter{\romannumeral1} demonstrates the performance of our model compared with the baseline and state-of-the-arts. The number of experts that achieve the best performance on test set are $\mathit{n}$=1 (98.12$\%$), $\mathit{n}$=2 (98.94$\%$), $\mathit{n}$=4 (99.12$\%$) and $\mathit{n}$=8 (99.60$\%$). From the results, we see that the models with CondConv ($\mathit{n}>1$) consistently perform better than the counterparts without CondConv ( $\mathit{n}$=1). As a reference, the baseline has 98.12$\%$ accuracy at the cost of 30.01 MFLOPs. The CondConv model has 99.60$\%$ classification accuracy with a computation complexity of 31.69 MFLOPs. There is an improvement of 1.48$\%$ in accuracy with a very small increase in FLOPs. To the best of our knowledge, the best performance on the dataset was 98.82$\%$ using CNN with local loss (Teng  \emph{et~al.} \cite{TengQi}). The second best result was 98.20$\%$ using a temporal convolution on the
spectrogram domain of the time series signal (Ravi \emph{et~al.} \cite{ravi2016deep}). Our result with CondConv is best reported, which surpasses the state-of-the-art results. The results imply that the proposed model demonstrates state-of-the-art performance using CNN while requiring almost the same computation cost.

\begin{table}[h]
	\caption{Performance on \textbf{WISDM} Dataset with  Different nums of Experts}
	\centering
	\begin{tabular}{ccccc}
		\toprule 
		\textbf{Model}&\textbf{Test Acc}&\textbf{FLOPs}\\
		\midrule
		CondConv(with $\mathit{n}$=1)& 98.12$\%$&30.01M\\
		CondConv(with $\mathit{n}$=2)& 98.94$\%$&30.25M\\
		CondConv(with $\mathit{n}$=4)& 99.12$\%$&30.73M\\
		CondConv(with $\mathit{n}$=8)&\textbf{99.60$\%$}&31.69M\\
		\midrule
		Ignatov \emph{et~al.}2018 \cite{ignatov2018real}  &93.32$\%$\\
		Teng \emph{et~al.}2020 \cite{TengQi}   &98.82$\%$\\
		Ravi \emph{et~al.}2016 \cite{ravi2016deep} &98.20$\%$\\
		\bottomrule
		\label{Tab1 }
	\end{tabular}
\end{table}

\subsubsection{$\textbf{The PAMAP2 dataset }${\color{black}}\cite{reiss2012introducing}}	
\indent The physical activity monitoring dataset is an open source dataset available at UCI repository, which contains extensive physical activities: both everyday household and sports performed by 9 participants wearing 3 inertial measurement units (IMUs) and a heart rate monitor. The IMU sensors were placed over the chest, wrist and side’s ankle on the dominant. The participants were asked to perform 12 protocol activities such as stand, sit, ascend stairs, descend stairs, rope jumping and run. In addition, some of them performed 6 optional activities such as watching TV, car driving, house cleaning and playing soccer. The sampling rate of heart rate monitor is 9Hz, and the sampling rate of IMUs is 100Hz; i.e. data is recorded 100 times per second. For the use of HAR, we subsample the IMU signals from 100Hz to 33.3Hz.\\
\indent As indicated, HAR is typically computed over a sliding window. The sliding window length is usually fixed. Different window lengths are selected by authors in various studies. To compare the result with other works, we selecte window size of 512 (5.12 seconds) to slide one instance at a time, which leads to a 78$\%$ overlap with around 473k samples. All samples are normalized into zero mean and unit variance. We randomly select 70$\%$ of the data in each class for training, the rest for test. The shorthand description of the baseline CNN is described as C(64)-C(128)-C(256)-FC-Sm, which consists of three convolutional layers and one fully connected layer. BN is applied before ReLU activation.
The batch size is set to 204 and Adam optimization \cite{kingma2014adam} is used for training. The learning rate is set to 0.001, 0.0005 and 0.00001 during 12.5$\%$, 25$\%$ and 62.5$\%$ of the total training time. 
\begin{figure}[htbp]
	\hspace*{3cm}\\
	\centering
	\includegraphics[width=7.3cm,height=5.5cm]{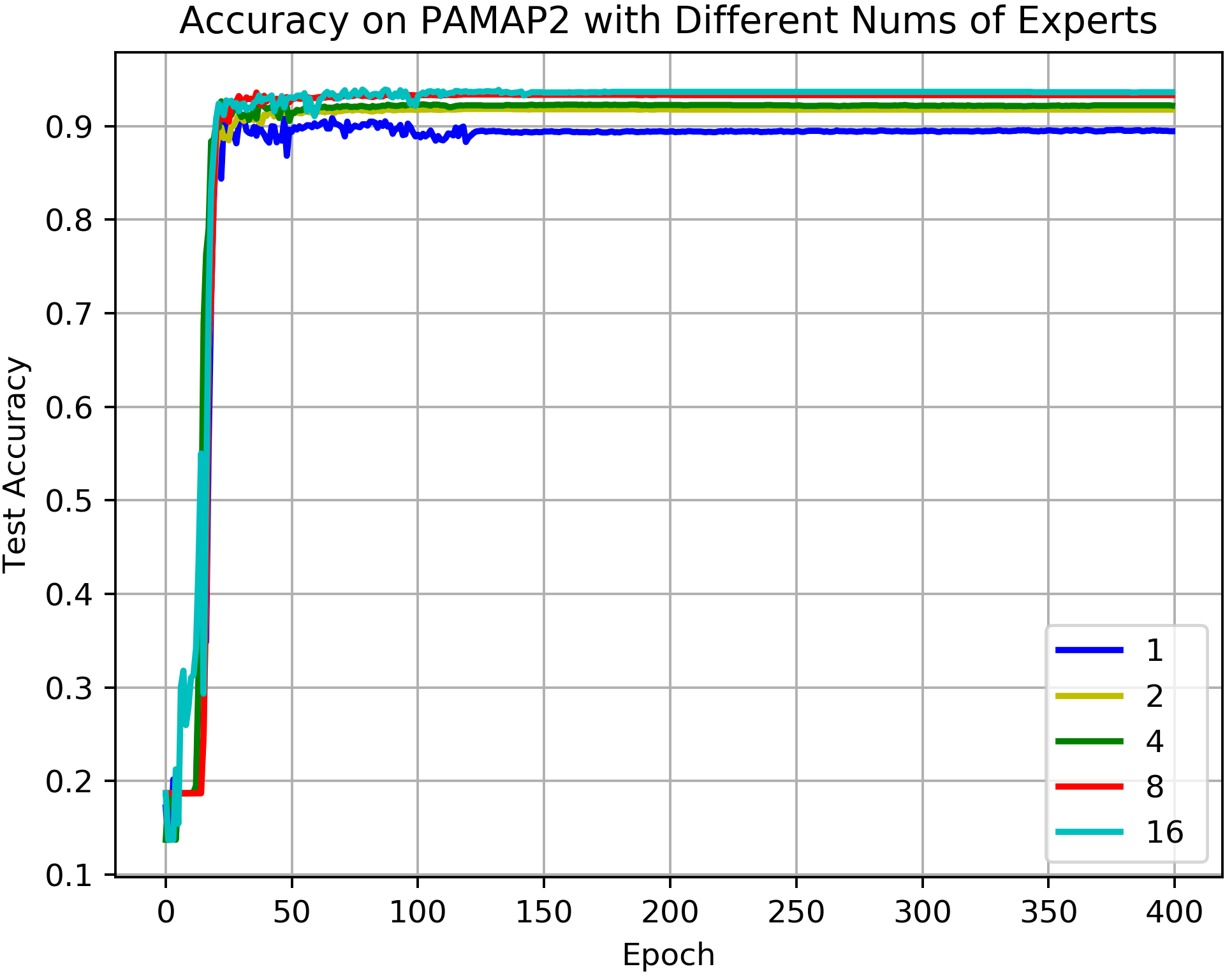}
	\caption{\label{Fig3 } Accuracy on \textbf{Pamap2} dataset with different nums of experts}
\end{figure}

\indent  Keeping all hyper-parameters except the number of experts identical, we train the best performing model using CondConv to see if it could improve the result further. Fig.3 shows the effect of increasing number of experts on the performance using the CondConv architectures with $\mathit{n}$=1,$\mathit{n}$=2,$\mathit{n}$=4,$\mathit{n}$=8 and $\mathit{n}$=16. It can be seen that the model performance consistently increases when the number of experts is greater than 1. Under a variety of $\mathit{n}$, we compare classification accuracy and FLOPs with the baseline of $\mathit{n}$=1, as well as the state-of-the-arts on the dataset. From the results in Table \uppercase\expandafter{\romannumeral2}, the number of experts that achieves the best results on test set are $\mathit{n}$=1(89.97$\%$), $\mathit{n}$=2(91.8$\%$), $\mathit{n}$=4(92.7$\%$), $\mathit{n}$=8(93.79$\%$) and $\mathit{n}$=16(94.01$\%$). Our method using CondConv with $\mathit{n}$=16 surpasses the baseline by 4.04$\%$, accompanied by a very small increase in computation cost. As can be seen in Table \uppercase\expandafter{\romannumeral2}, the best published result on this dataset using CNN is to our knowledge 91.4$\%$(Yang \emph{et~al.} \cite{yang2018dfternet}). The proposed method surpasses the state-of-the-art result by a large margin. This implies that we can exploit this CondConv as a drop-in replacement of standard convolutions to achieve better results with only a small increase in computation cost. 

\begin{table}[h]
	\caption{Performance on \textbf{PAMAP2} Dataset with  Different nums of Experts}
	\centering
	\begin{tabular}{ccc}
		\toprule 
		\textbf{Model}&\textbf{Test Acc}&\textbf{FLOPs}\\
		\midrule
		CondConv(with  $\mathit{n}$=1)& 89.97$\%$&31.57M\\
		CondConv(with  $\mathit{n}$=2)& 91.80$\%$&31.81M\\
		CondConv(with  $\mathit{n}$=4)& 92.70$\%$&32.31M\\
		CondConv(with  $\mathit{n}$=8)&93.79$\%$&33.25M\\
		CondConv(with  $\mathit{n}$=16)&\textbf{94.01$\%$}&35.17M\\
		\midrule
		Yang \emph{et~al.}2018  \cite{yang2018dfternet}   &91.40$\%$\\
		Zeng \emph{et~al.}2018  \cite{zeng2018understanding}    &89.96$\%$\\
		Khan \emph{et~al.}2016 \cite{khan2016optimising} &86.00$\%$\\

		\bottomrule
		\label{Tab2 }
	\end{tabular}
\end{table}

\subsubsection{$\textbf{The UNIMIB-SHAR dataset  }${\color{black}} \cite{micucci2017unimib}}	
\indent UNIMiB-SHAR is a new dataset including 11771 samples designed for the use of HAR and fall detection. In a supervised condition, the 30 subjects of ages ranging from 18 to 60 years wearing a Samsung Galaxy Nexus I9250 smartphone were instructed to perform activities.  Each activity was performed 2 or 6 times. The half of all participants placed the smartphone in their left pocket, and the other half in their right pocket. An embedded Bosh BMA220 3D accelerometer was used to generate examples. The whole dataset is composed of 17 fine grained classes which is further grouped into two coarse grained classes: one containing samples of 9 types of activities of daily living(ADLs) and the other containing samples of 8 types of falls. 

For fair comparison, the sliding window with a fixed length T=151 is selected, which equals to approximately 3s. Since the accelerometer signals are recorded at a constant sampling rate of 50 Hz, for each activity, the accelerometer signal is comprised of 3 vectors of 151 values, one for each acceleration direction. Thus, the whole dataset contains 11,771 windows of size 151*3 in total, which describes both ADLs (7759) and falls (4192) unequally distributed across activity types. The architecture of the baseline CNN is C(128)-C(256)-C(384)-FC-Sm, which contains three convolutional layers and one fully connected layer. At the last, we use a 1x1 CondConv layer to replace the final fully connected classification layer. The samples are split into 70$\%$ training and 30$\%$ test set. Adam optimizer \cite{kingma2014adam} is used to train with batch size of 203. The learning rate is set to 0.0004, 0.00001 and 0.000001 during 12.5$\%$, 25$\%$ and 62.5$\%$ of the total training time.

\begin{figure}[htbp]
	\hspace*{3cm}\\
	\centering
	\includegraphics[width=7.7cm,height=5.4cm]{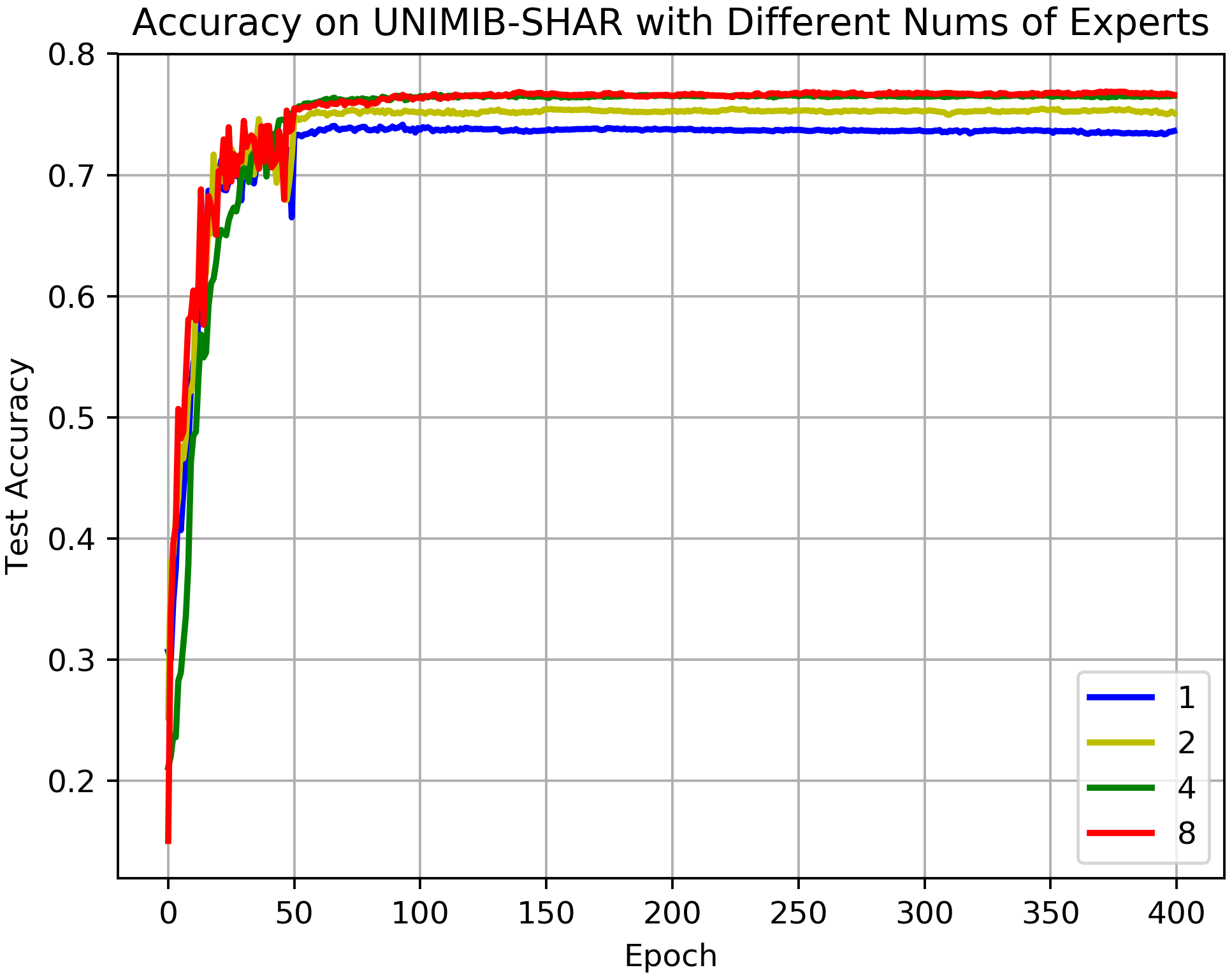}
	\caption{\label{Fig4}  Accuracy on \textbf{UNIMIB-SHAR} dataset with different nums of experts.}
\end{figure}

\indent We evaluate the performance of our proposed method with various number of experts on the dataset. Fig.4 shows that the accuracy will increase as the number of experts grows, which is consistent with our motivation. The CNN that utilizes CondConv always performs better than that without CondConv. Table III demonstrates the performance of our model compared with the baseline and other state-of-the-arts in terms of accuracy and FLOPs. It can be seen that our method achieves 1.32$\%$, 2.74$\%$ and 3.16$\%$ improvements over baseline with $\mathit{n}$=2, $\mathit{n}$=4 and $\mathit{n}$=8 respectively for this task. There is only a small increase in computation cost. In addition, our model using CondConv outperforms other state-of-the-arts. When compared to the best result obtained by Li \emph{et~al.} \cite{li2018comparison} using CNN, our method with $\mathit{n}$=8 achieves 2.34$\%$ improvement. Our CondConv also  surpasses the Long \emph{et~al's} method \cite{long2019dual} by 1.28$\%$, which uses dual residual
networks. Under same parameter configurations, by increasing the number of experts, the CondConv with sufficient regulation is able to enhance the expression ability of CNN by a large margin. 
\begin{table}[htbp]
	\caption{Performance of Different Experts for \textbf{UNIMIB-SHAR} dataset}
	\centering
	\begin{tabular}{ccc}
		\toprule 
		\textbf{Model}&\textbf{Test Acc}&\textbf{FLOPs}\\
		\midrule
		CondConv(with $\mathit{n}$=1)& 74.15$\%$&62.26M\\
		CondConv(with $\mathit{n}$=2)& 75.47$\%$&62.71M\\
		CondConv(with $\mathit{n}$=4)& 76.89$\%$&63.14M\\
		CondConv(with $\mathit{n}$=8)&\textbf{77.31$\%$}&64.26M\\
	
		\midrule
		 Li \emph{et~al.}2018 \cite{li2018comparison}&74.97$\%$\\
		 Long \emph{et~al.}2019 \cite{long2019dual}&76.03$\%$\\
		\bottomrule
		\label{Tab3}
	\end{tabular}
\end{table}
\subsubsection{$\textbf{The OPPORTUNITY dataset  }${\color{black}} \cite{chavarriaga2013opportunity}}	
\indent The OPPORTUNITY dataset is publicly available on the UCI Machine Learning repository, which comprises both static/periodic and sporadic activities collected with sensors of different modalities integrated into the environment and on the subjects, in a daily living scenario. The samples were recorded from four subjects performing morning activities, in which each subject was asked to perform one ADL session and one drill session. During the ADL session, without any strict restriction, subjects performed a session five times with activities such as preparing and drinking a coffee, preparing and eating a sandwich, cleaning up, and so on. During the drill session, subjects were instructed to perform 20 repetitions of a predefined sorted set of 17 activities. 

\begin{figure}[htbp]
	\hspace*{3cm}

	\centering
	\includegraphics[width=7.7cm,height=5.4cm]{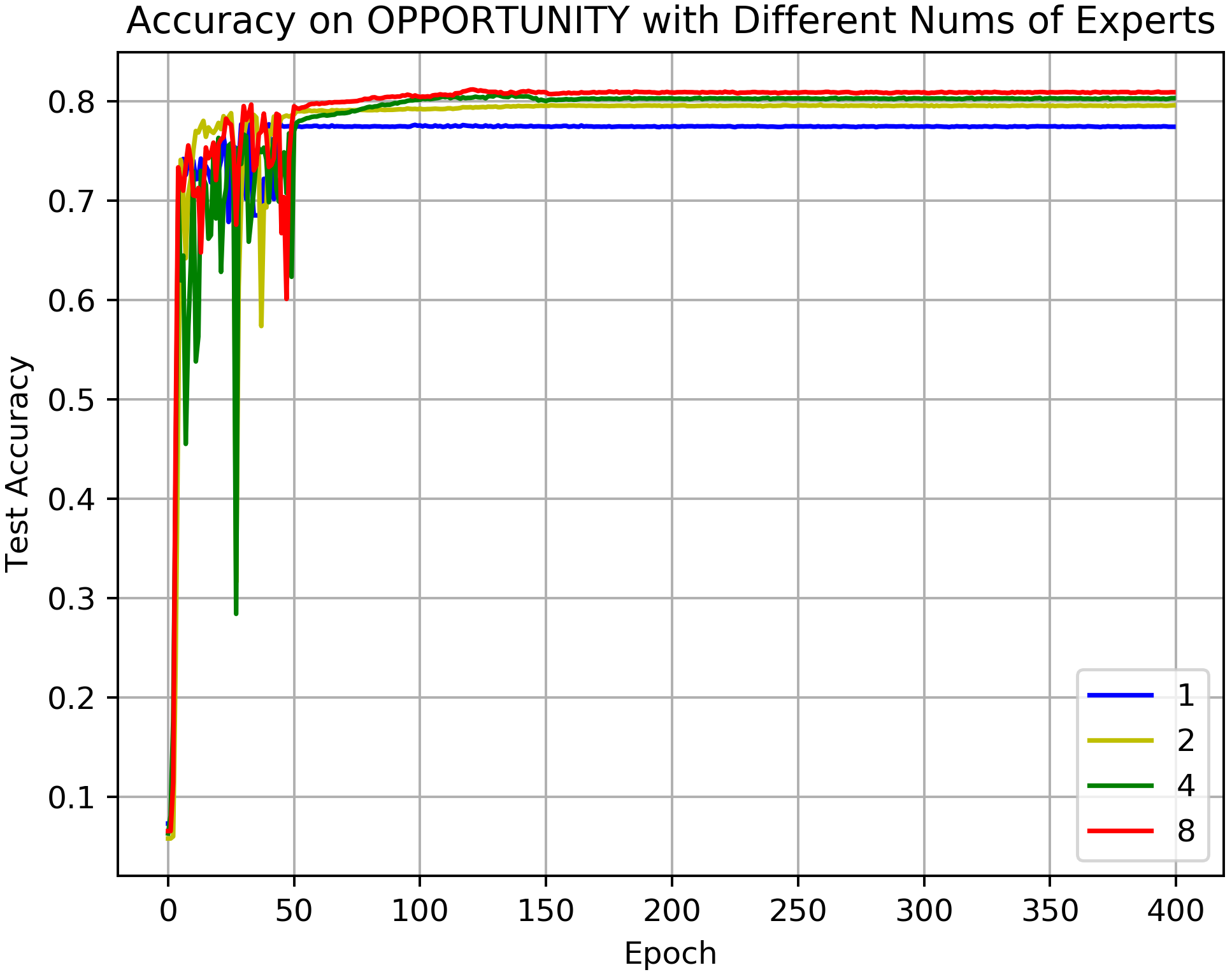}
	\caption{\label{Fig5}   Accuracy on \textbf{OPPORTUNITY} dataset with different nums of experts. }
\end{figure}

\indent The dataset has been used in numerous activity recognition challenges. In this paper, we evaluate our method on the same subset employed in previous OPPORTUNITY challenge, which contains the samples collected from 4 subjects with only on-body sensors. The sensor signals are recorded at a sampling rate of 30Hz from 12 locations on the dominant and annotated with 18 mid-level gesture annotations. ADL1, ADL2 and ADL3 from subject 1, 2 and 3 are used as training set. ADL4 and ADL5 from subject 4 and 5 are used as test set. The size of sliding time window and sliding step length are set to 64 and 8 respectively, which generates approximately 650k samples. The baseline model is a deep CNN, whose shorthand description is presented as C(64)-C(64)-C(128)-C(128)-C(256)-Fc-Sm, that contains five convolutional layers and one fully connected layer. For this experiment, The initial learning rate is set as 0.0001, which will be reduced by a factor of 0.1 after each 50 epochs using Adam with default parameters. Initial batch size is set to 204.\\ 

\indent We characterize the effect of the number of experts employed to increase model capacity or performance. Fig.5 shows that with sufficient regulation increasing the number of experts tends to increase the performance on the OPPORTUNITY dataset. The results of the proposed method are shown in Table \uppercase\expandafter{\romannumeral4}, which also includes a comprehensive list of past published deep learning techniques employed on the dataset. Among deep architectures, the CondConv systematically performs best on the dataset. It can be seen that the results of the baseline CNN are close to those obtained previously by Zeng \emph{et~al.} \cite{zeng2014convolutional} using CNN on raw signal data. The CondConv with the number of experts greater than 1 consistently outperforms our baseline, with a very small increase in computation cost. We also reproduce the state-of-the-art DeepConvLSTM using our experiment setup. The CondConv systematically performs better
than the DeepConvLSTM, improving the performance by 2.28$\%$ on average on the OPPORTUNITY dataset.

\begin{table}[h]
	\caption{Performance of Different Experts for \textbf{OPPORTUNITY} dataset}
	\centering
	\begin{tabular}{ccc}
		\toprule 
		\textbf{Model}&\textbf{Test Acc}&\textbf{FLOPs}\\
		\midrule
		CondConv(with $\mathit{n}$=1)& 77.5$\%$&114.23M\\
		CondConv(with $\mathit{n}$=2)& 78.7$\%$&115.57M\\
		CondConv(with $\mathit{n}$=4)& 80.9$\%$&117.43M\\
		CondConv(with $\mathit{n}$=8)&\textbf{81.18$\%$}&121.56M\\
		
		\midrule
		Zeng \emph{et~al.}2014 \cite{zeng2014convolutional}&76.83$\%$\\
		Ordóñez   \emph{et~al.}2016 \cite{article}   &78.90$\%$\\		
		Hammerla   \emph{et~al.}2016 \cite{hammerla2016deep}   &74.50$\%$\\
		\bottomrule
		\label{Tab1}
	\end{tabular}
\end{table}

\subsection{The Ablation studies }
\indent To better understand model design with the CondConv block, we conduct several ablation studies to further explore why the CondConv with larger model capacity is able to improve accuracy while maintaining efficient inference. Our ablation experiments are performed on the UNIMIB-SHAR dataset, and all hyper-parameters are exactly the same as used above. Finally, we also evaluate the actual inference time of our model on an Android smartphone.\\
\begin{figure}[htbp]	
	
	\begin{minipage}{0.5\linewidth}
		\centering
		\vspace{-0.6cm}
		\hspace*{0.7cm}
		\includegraphics[width=7cm,height=4.5cm]{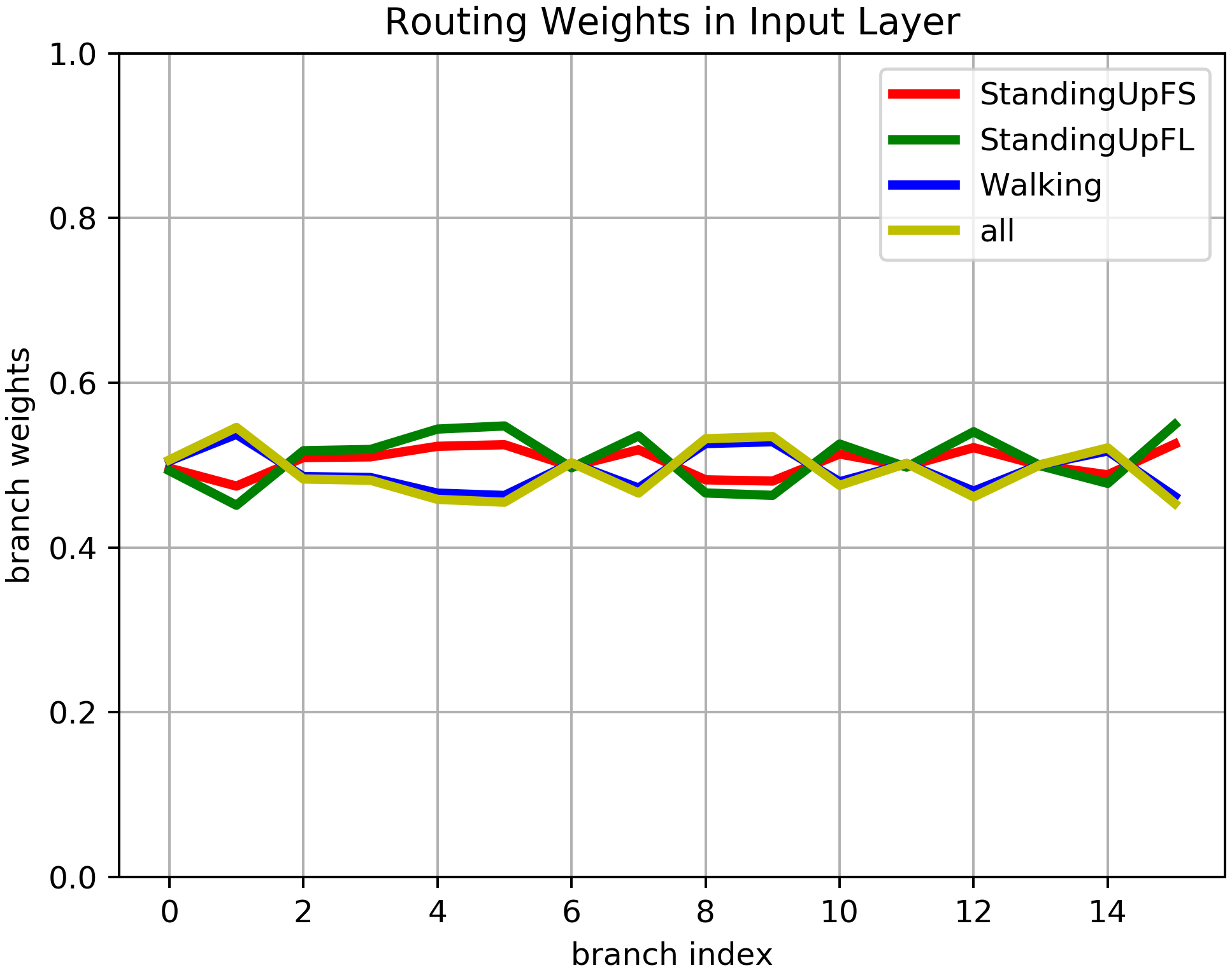}	 	
		\centering\hspace*{-4.5cm}
	\end{minipage}

	\begin{minipage}{0.5\linewidth}
		{\centering\hspace*{0.8cm}\includegraphics[width=7cm,height=4.5cm]{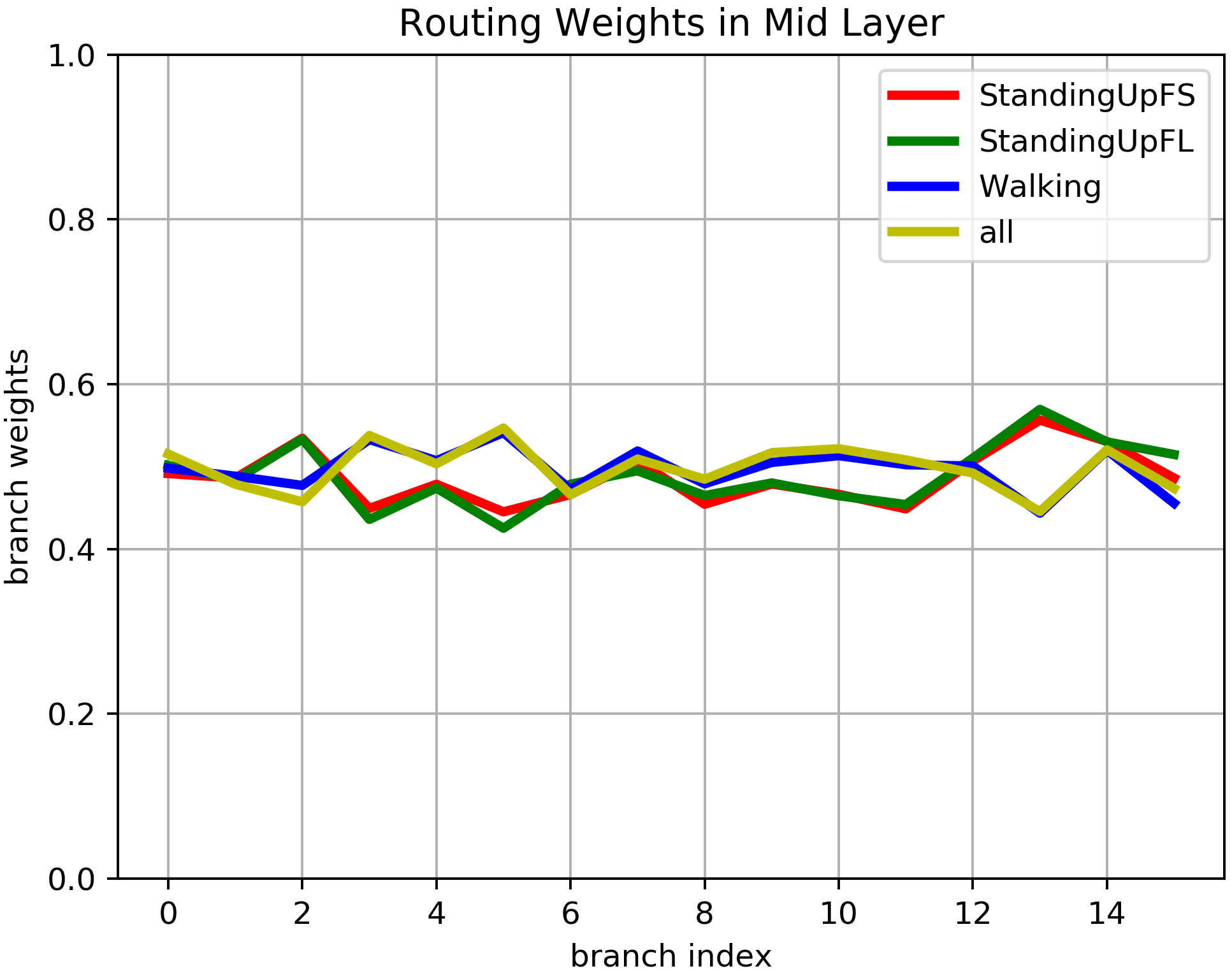}}
		\centering\hspace*{-4.5cm}			
	\end{minipage}
	
	\begin{minipage}{0.5\linewidth}
		{\centering\hspace*{0.8cm}\includegraphics[width=7cm,height=4.5cm]{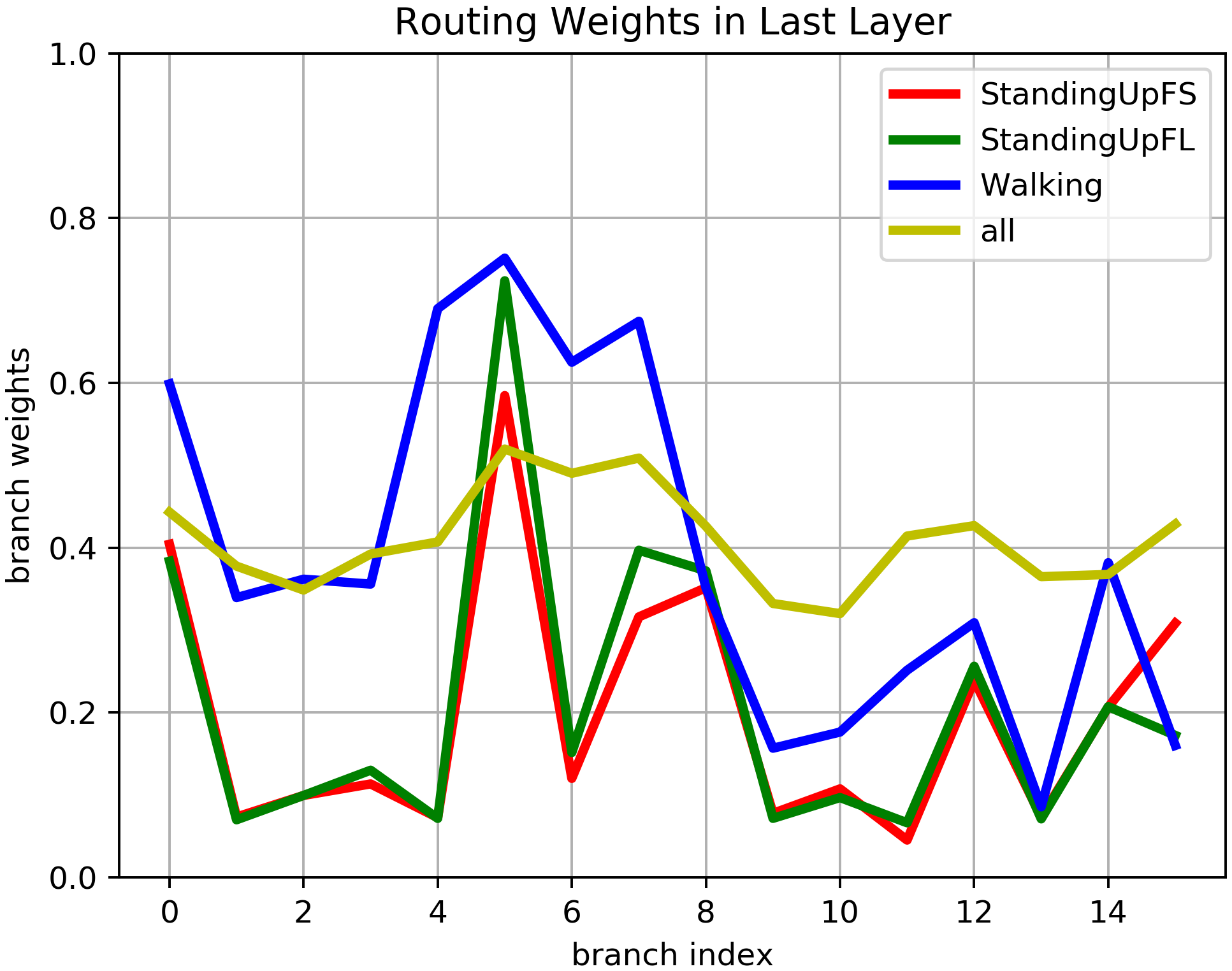}}		 	
		\centering\hspace*{-4.6cm}
	\end{minipage}
	
	\caption{\label{Fig6}  Mean routing weights for three classes across the \textbf{UNIMIB-SHAR} dataset at three different depths in our model}
\end{figure}

\indent First, we study the influence of routing weights across different classes of activities at three different depths in the network. As mentioned above, if all the experts have the same routing weight for each example, the CondConv will degenerate into standard convolutions. Thus, all the experts are example-dependent, and each individual example can yield different activation weights. We apply the CondConv in all convolutional layers as well as the final fully connected classification layer. Results are shown in Fig.6. It can be seen that the value discrepancy is increased layer by layer. For shallow layers, the distributions of routing weights of different experts are very close across classes, while in deep layers they are diverse. That is to say, the experts are more class specific or sensitive to high-level features, which suggests that there is no significant performance improvement if the CondConv layer is applied near the input of the network. In particular, we also find that the examples from the similar activities such as StandingUpFL and StandingUpFS tend to follow very close distribution.\\

\indent Next, to demonstrate the superiority of our method, we use the CondConv to compute the confusion matrices on the UNIMIB-SHAR dataset. As can be seen in Fig.7, for the similar activities such as StandingUpFL and StandingUpFS, the baseline CNN made 31 errors, while the CondConv in case of $\mathit{n}=8$ misclassified only 17 activities. Though the experts activated by the similar activities follow almost the same distribution, their combination is still able to offer better results, which indicates that multiple experts are often more useful than one. The CondConv is able to enhance the expression ability of CNN by a large margin via increasing the number of experts.\\
	\begin{figure}[htbp]

	\begin{minipage}{0.4\linewidth}
		\centering
		\vspace*{-0.6cm}
		\hspace{-0.4cm}
		\includegraphics[width=9cm,height=7.6cm]{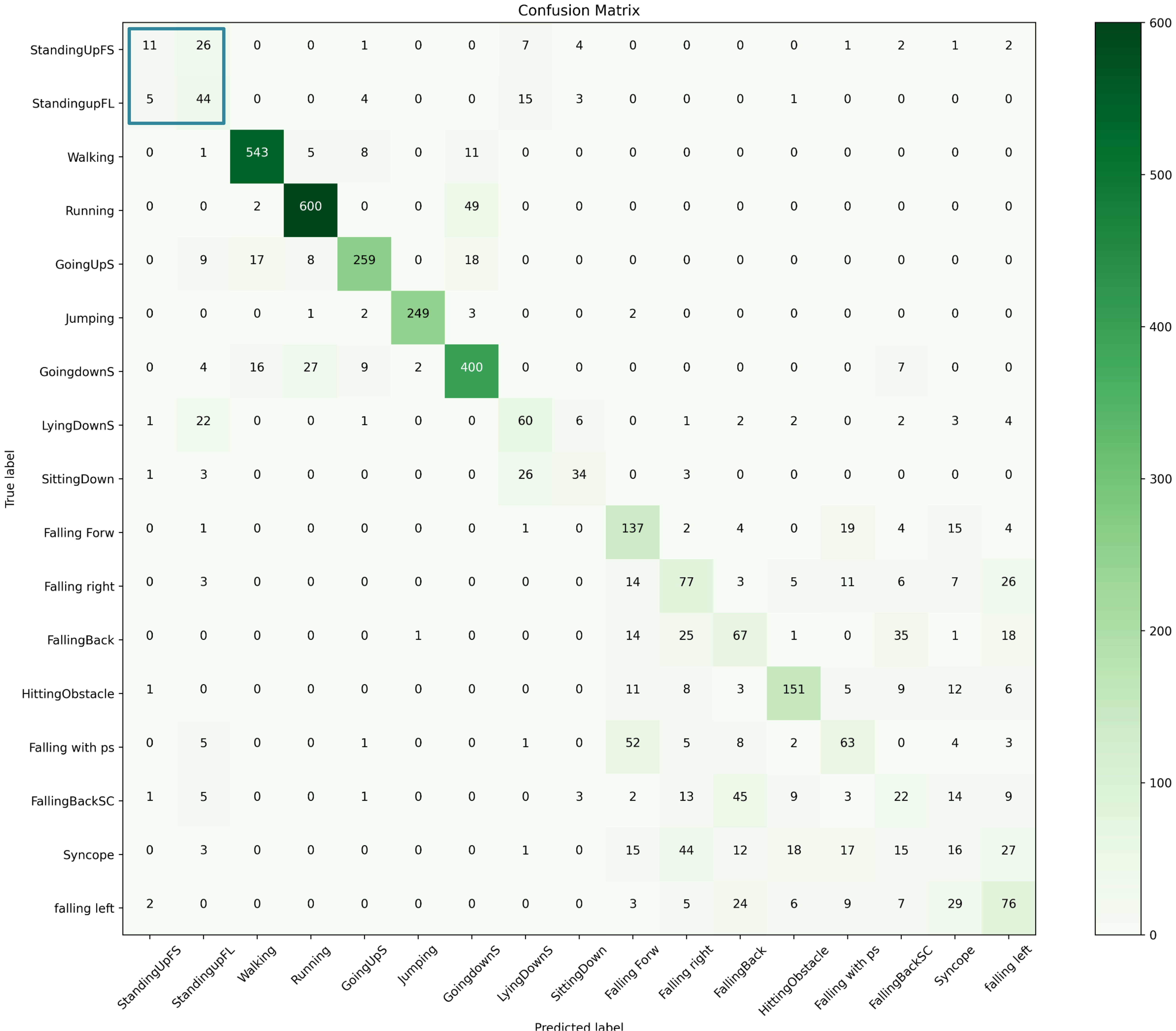}	 	
		\centering\hspace*{7cm}
	\end{minipage}

	\begin{minipage}{0.4\linewidth}
		\hspace{-0.4cm}
		{\centering\includegraphics[width=9cm,height=7.6cm]{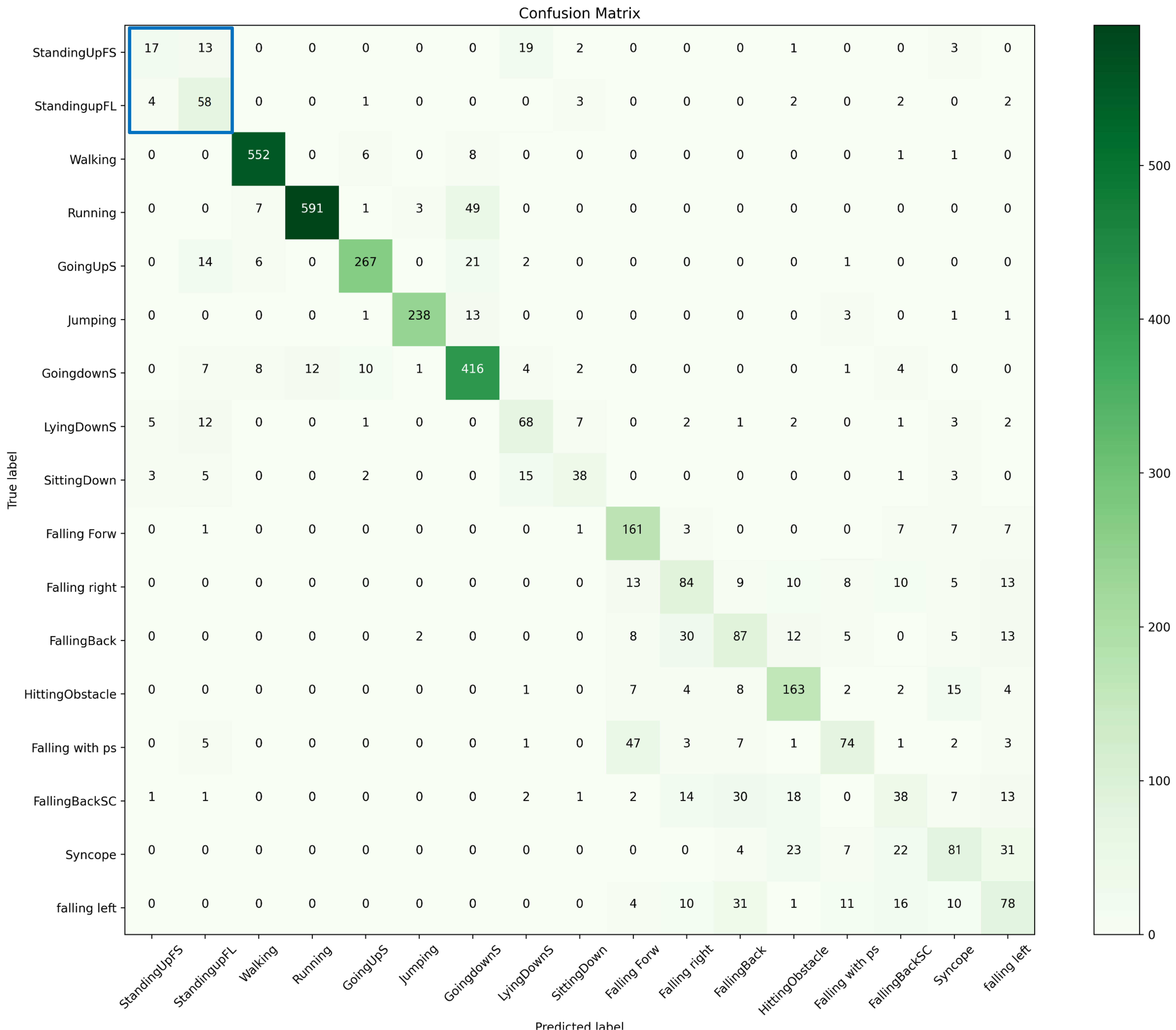}}
		\centering\hspace*{5cm}			
	\end{minipage}

	\caption{\label{Fig7} Confusion matrix for UNIMIB-SHAR dataset using the CondConv with $\mathit{n}$=1 and $\mathit{n}$=8 from top to bottom.}
\end{figure}

\indent Next, we evaluate the distribution of routing weights activated by all the examples in the UNIMIB-SHAR test set in the final CondConv layer. The main purpose of this evaluation is to disentangle the influence of different experts at deeper layers. Fig.8 shows the routing weights follow a bi-modal distribution, and most of them approximately equal to 0 or 1. Without using any L1 regularization technique, most experts are sparsely activated. That is to say, for each individual example, only a small portion of the entire network is activated, which suggests an explanation why the CondConv is able to realize efficient inference with larger model capacity.\\

\begin{figure}[htbp]
	\vspace{-0.2cm} 
	\centering
	
	\includegraphics[width=8cm,height=5.3cm]{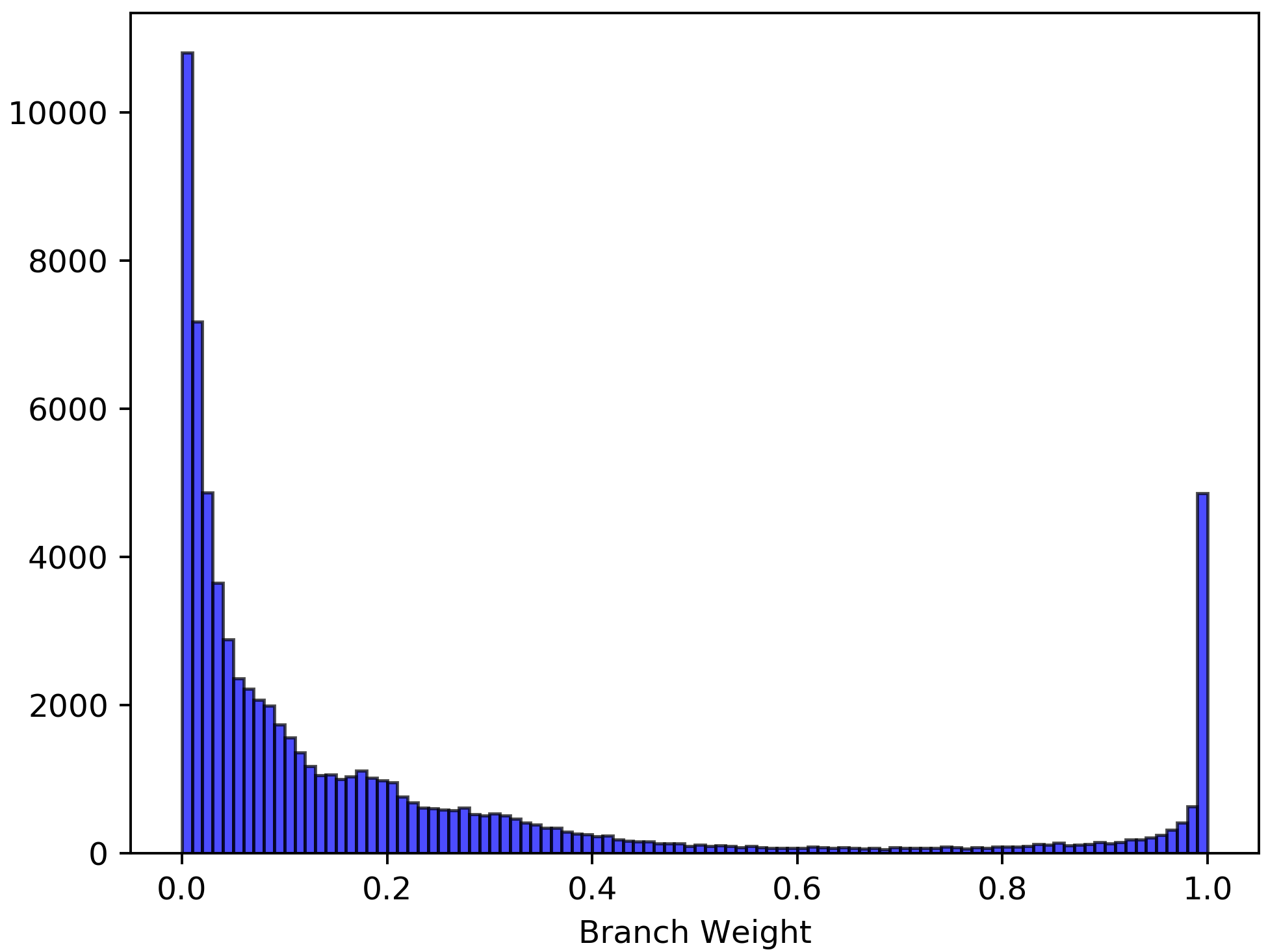}
	\caption{\label{Fi8}  Distribution of routing weights in the final CondConv layer. Routing weights follow a bi-modal distribution.}
\end{figure}

\indent We then study the variation of routing weights within one class in the final CondConv layer. Results are shown in Fig.9. We find that even within one class the routing weights between examples show much higher variance. In addition, to gain a better understanding of experts in the final CondConv layer, we visualize several typical activity examples of top 4 classes with highest activated values on 8 difference experts, as shown in Fig.10.\\

\begin{figure}[htbp]
	\hspace*{0.2cm}\\
	\vspace{-0.8cm} 
	\centering
	\includegraphics[width=9cm,height=3.5cm]{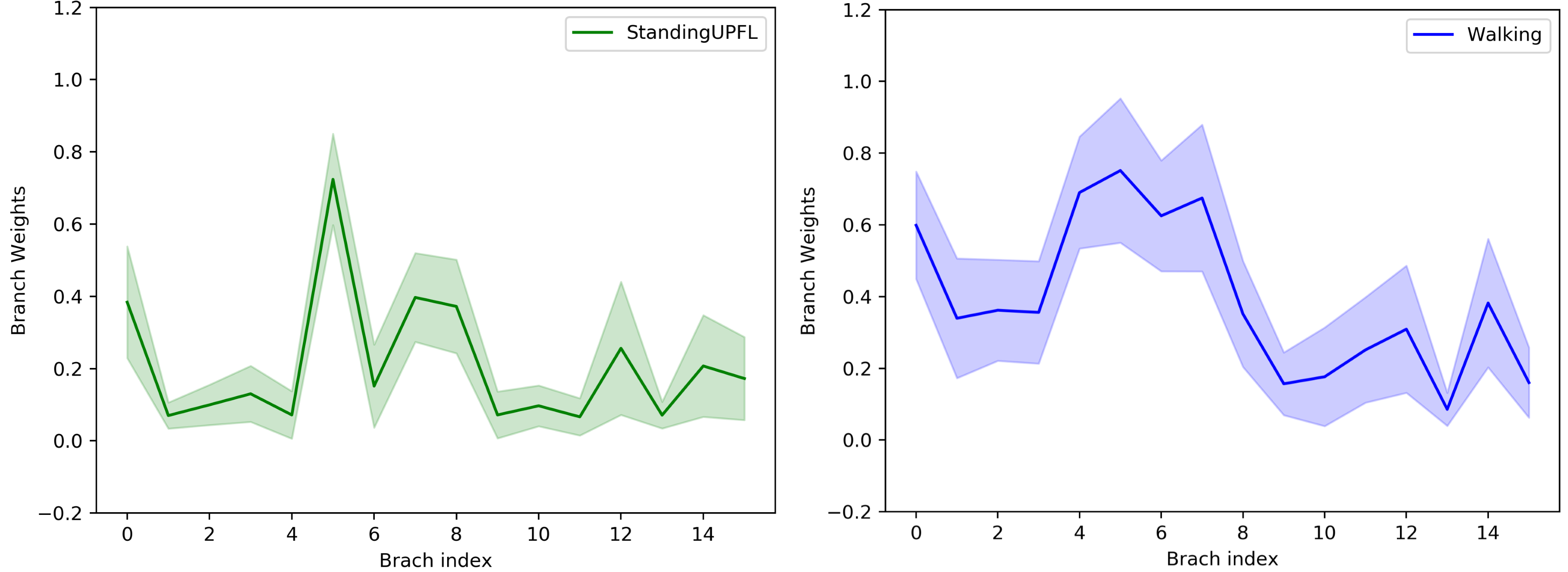}
	\caption{\label{Fig9}  Routing weights in the final CondConv layer in our model for 2 classes averaged across UNIMIB-SHAR test set. Error bars indicate one standard deviation.}
\end{figure}

\begin{figure}[htbp]
	\hspace*{2cm}\\
	\vspace{-1.1cm} 
	\centering
	
	\includegraphics[width=8.9cm,height=5.3cm]{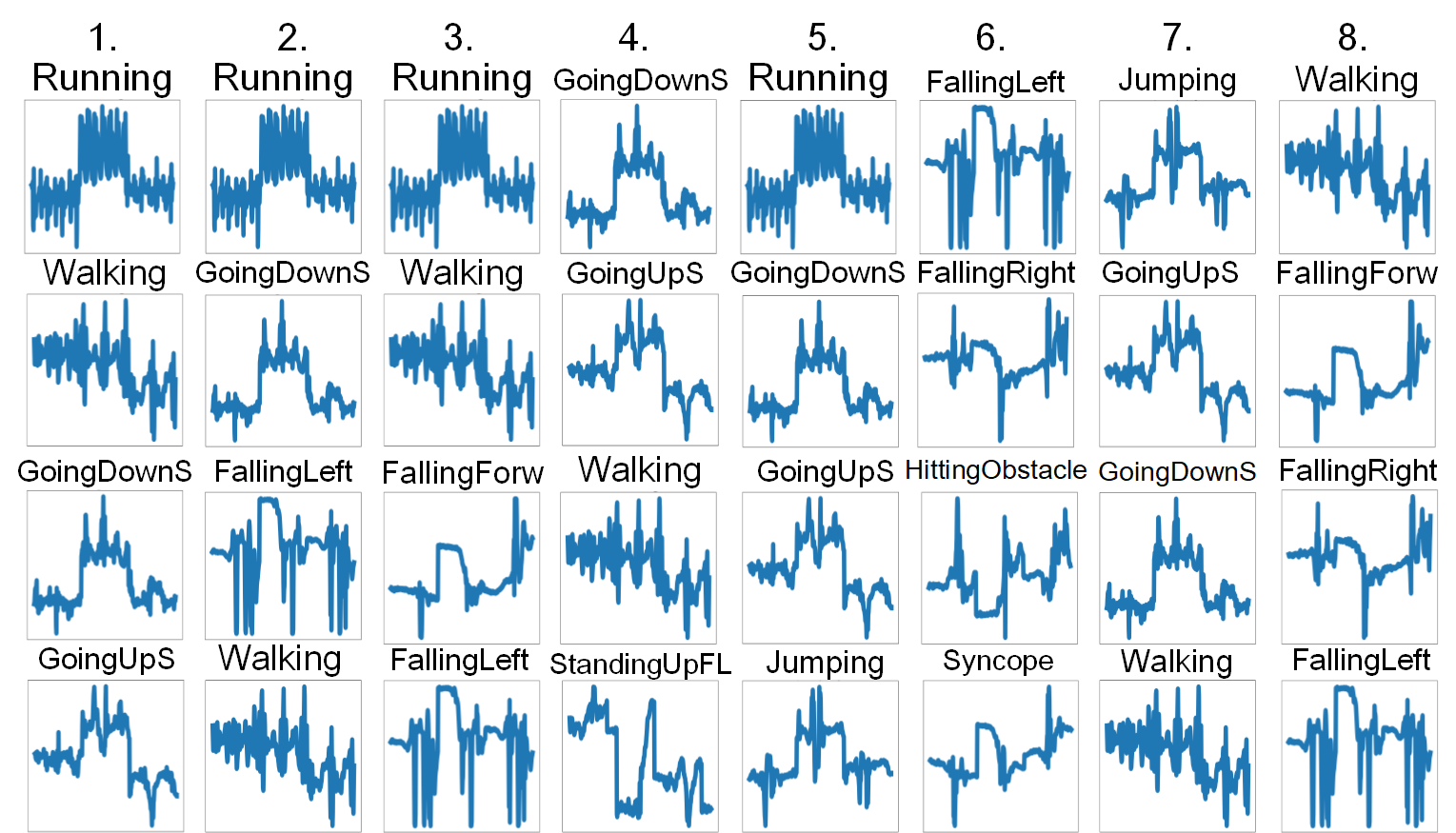}
	\caption{\label{Fig10}  In this figure, from top to bottom, we could see that most experts are activatied by Running due to the imbalanced dataset in which Running accounts for a large fraction of all 17 categories. The fourth expert is more specific to GoingUp and GoingDown and the sixth expert is most activated by FallingLeft and FallingRight.}
\end{figure}

\indent Finally, we evaluate the actual inference time of the CondConv models on a smartphone. The open source APP introduced in \cite{singh2017human} is directly utilized for the evaluation, which is a smartphone-based application for mobile HAR. A screenshot of the APP's user interface is shown in Fig.11. The CondConv models with $\mathit{n}=1$ and $\mathit{n}=8$ are trained on the WISDM dataset.  We convert the models into .pb file, which are deployed to build an Android application. Our experiment is implemented on a Huawei Mate 30 device with the Android OS(10.0.0). As shown in Table V, it can be seen that the CondConv with $\mathit{n}=8$ has almost the same inference speed with baseline in the real implementation.

\begin{figure}[htbp]
	\hspace*{0.2cm}\\
	\vspace{0.5cm} 
	\centering
	\includegraphics[width=3.2cm,height=5.8cm]{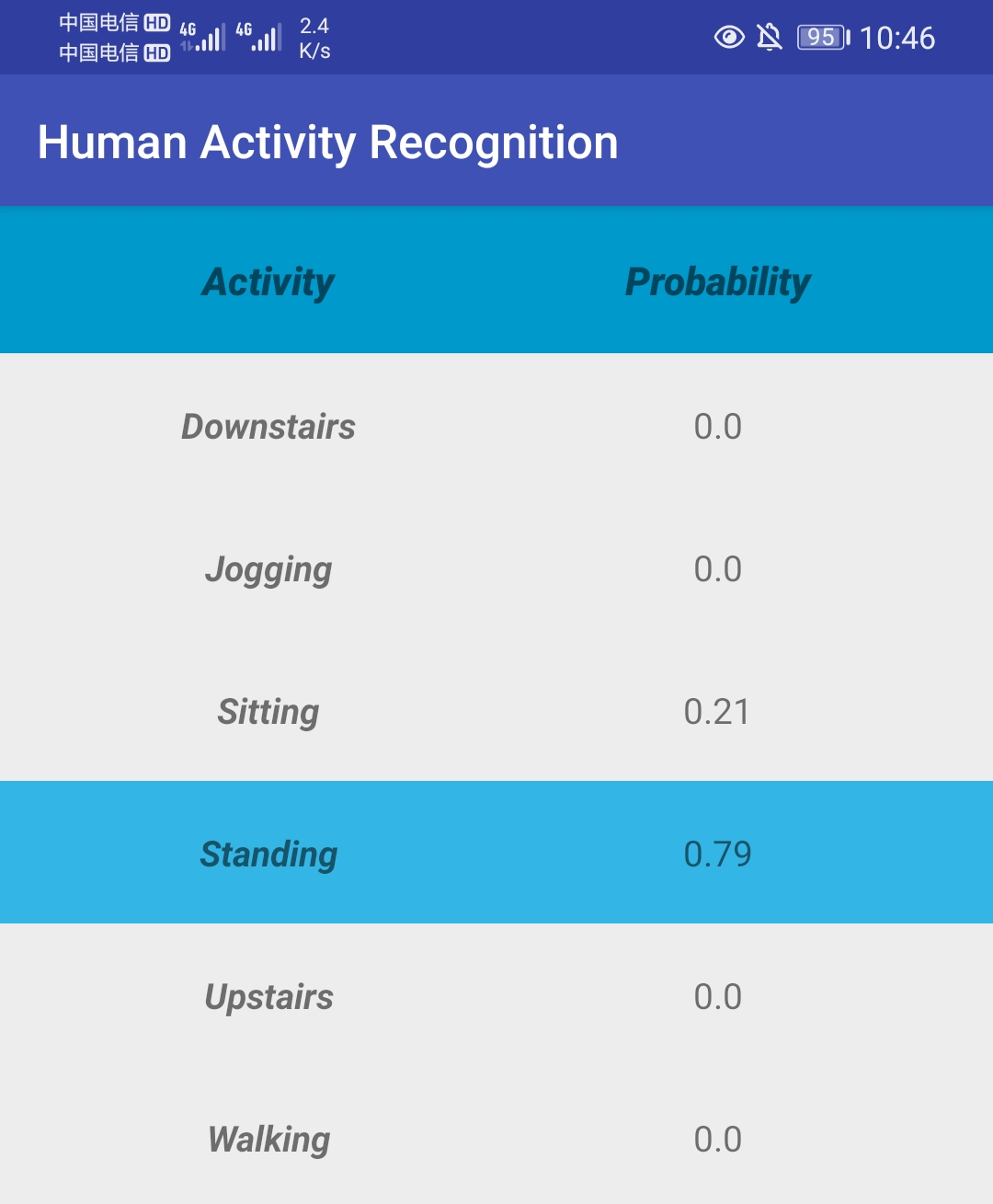}
	\caption{\label{Fig11} Screenshot of the APP's user interface}
\end{figure}

\begin{table}[h]
	\caption{Inference time between Conv and CondConv}
	\centering
	\begin{tabular}{cccccc}
		\toprule 
		\textbf{Model}&\textbf{Inference Time(ms/window)}\\
		\midrule
		CNN(Baseline)&228-272ms\\
		\midrule
		CondConv($\mathit{n}$=8)&241-292ms\\
	
		\bottomrule
		\label{Tab7}
	\end{tabular}
\end{table}	
\section{Conclusion}
\indent Recently, deep CNNs have achieved state-of-the-art performance on various mobile and wearable HAR tasks. However, this technique is severely hampered by the computation power in current mobile and wearable devices. A high number of computations in deep leaning increases computational time and is not suitable for real-time HAR on mobile and wearable devices. Shallow and conventional machine learning methods could not achieve good performance. Therefore, deep learning methods that can balance the trade-off between accuracy and computation cost is highly needed. In this paper, we have presented an efficient solution for HAR on mobile and wearable devices via replacing conventional convolutions with CondConv. The proposed CondConv method is evaluated in on four public HAR benchmark datasets, WISDM dataset, PAMAP2 dataset, UNIMIB-SHAR dataset, and OPPORTUNITY dataset, achieving state-of-the-art accuracy without compromising inference speed. We have also performed various ablation experiments to show how such a larger network is clearly preferable to the baseline while requiring a similar amount of operations. On the whole, with efficient regulation, the proposed method can greatly improve recognition accuracy of the existing HAR using CNN without compromising computation cost, which is very suitable for HAR that has strict latency constrains. By combining the efficient architecture design with any existing CNN based HAR method, we are able to perform real-time HAR tasks on mobile and wearable devices.\\

\end{document}